\documentclass[conference]{IEEEtran}
\IEEEoverridecommandlockouts
\usepackage{cite}
\usepackage{indentfirst}
\usepackage{float}
\usepackage{amsmath,amssymb,amsfonts}
\usepackage{graphicx}
\usepackage{textcomp}
\usepackage{xcolor}
\usepackage{booktabs}
\usepackage{multirow}
\usepackage{amsmath,amssymb,amsfonts}
\usepackage{algorithmic}
\usepackage{graphicx}
\usepackage{booktabs}
\usepackage{textcomp}
\usepackage{xcolor}
\def\BibTeX{{\rm B\kern-.05em{\sc i\kern-.025em b}\kern-.08em
    T\kern-.1667em\lower.7ex\hbox{E}\kern-.125emX}}
    
\begin{document}

\title{VFM-VLM: Vision Foundation Model and Vision Language Model based Visual Comparison for 3D Pose Estimation }

\author{
\IEEEauthorblockN{Md Selim Sarowar}
\IEEEauthorblockA{\textit{Dept. of Electronic Engineering} \\
\textit{Yeungnam University}\\
Advanced Visual Intelligence Lab\\
selim.sarowar12@gmail.com}
\and
\IEEEauthorblockN{Sungho Kim}
\IEEEauthorblockA{\textit{Dept. of Electronic Engineering} \\
\textit{Yeungnam University}\\
Advanced Visual Intelligence Lab\\
sunghokim@yu.ac.kr}
}

\maketitle

\begin{abstract}
Vision Foundation Models (VFMs) and Vision Language Models (VLMs) have revolutionized computer vision by providing rich semantic and geometric representations. This paper presents a comprehensive visual comparison between CLIP based and DINOv2 based approaches for 3D pose estimation in hand object grasping scenarios. We evaluate both models on the task of 6D object pose estimation and demonstrate their complementary strengths: CLIP excels in semantic understanding through language grounding, while DINOv2 provides superior dense geometric features. Through extensive experiments on benchmark datasets, we show that CLIP based methods achieve better semantic consistency, while DINOv2 based approaches demonstrate competitive performance with enhanced geometric precision. Our analysis provides insights for selecting appropriate vision models for robotic manipulation and grasping, picking applications.\cite{8996450}
\end{abstract}

\section{Introduction}

A paradigm shift from purely geometric methods to semantically aware systems is represented by the incorporation of Vision Foundation Models (VFMs) and Vision Language Models (VLMs) into 3D pose estimation. While mathematically correct, existing deep learning ways for 6D pose estimation\cite{tyree20226dofposeestimationhousehold} lack the contextual knowledge required for practical robotic manipulation and human computer interaction.

Two prominent paradigms have emerged from recent developments in self-supervised learning: DINOv2 \cite{oquab2023dinov2}, which uses self-distillation to capture rich visual features without explicit language supervision, and CLIP \cite{Radford2021LearningTV}, which learns joint vision language representations through contrastive learning. Although both methods have demonstrated impressive performance in a variety of computer vision applications, their use in 3D pose estimation is still largely unexplored.

This paper makes the following contributions:
\begin{itemize}
\item A systematic comparison of CLIP based and DINOv2 based architectures for 6D object pose estimation in grasping scenarios
\item Visual analysis of prediction quality through 2D keypoint projection and 3D bounding box visualization
\item Quantitative evaluation using standard metrics (ADD, ADD-S, rotation error, translation error)
\item Insights between semantic understanding and geometric precision
\end{itemize}

\section{Related Work}

\subsection{Taxonomy of Pose Estimation}

3D pose estimation can be systematically categorized along multiple dimensions, forming a comprehensive taxonomy that guides our understanding of the field's evolution.

\textbf{Input Modality:} Methods are divided into RGB-only approaches, which leverage color information and learned representations, RGB-D methods that exploit depth sensors for geometric constraints, and multi-modal approaches combining various sensor inputs. Recent VFM based methods primarily operate on RGB input but extract rich geometric and semantic features.

\textbf{Output Representation:} Direct pose regression methods predict 6D pose parameters (rotation and translation)\cite{11094705} end-to-end. Keypoint-based approaches first detect 2D projections of 3D keypoints, then solve PnP for pose recovery. Dense correspondence methods\cite{haugaard2022surfembdensecontinuouscorrespondence} establish pixel-to-model mappings before pose optimization. Hybrid methods combine multiple representations for robustness.

\textbf{Learning Paradigm:} Supervised methods require extensive labeled pose annotations. Self-supervised approaches leverage geometric constraints (epipolar geometry, photometric consistency)\cite{10447716} or reconstruction losses. Foundation models introduce a new paradigm of pre-training on massive unlabeled datasets followed by task specific fine-tuning, bridging the gap between pure supervision and self-supervision.

\textbf{Architecture:} CNN based methods (PoseCNN, DenseFusion)\cite{xiang2018posecnn} \cite{10923719} dominated early deep learning approaches. Transformer-based architectures capture global context and long range dependencies. Point cloud networks (PointNet, PointNet++)\cite{qi2017pointnetdeeplearningpoint} process 3D geometric data directly. VFM based approaches leverage pre-trained vision transformers with billions of parameters, representing the current frontier.

This taxonomy reveals that VFM aided methods represent a convergence point: they primarily use RGB input, produce multiple output representations (keypoints, dense features, direct pose), leverage self-supervised pre-training at scale, and employ transformer architectures. Our work positions CLIP and DINOv2 within this taxonomy as representatives of language grounded and pure visual self-supervised paradigms, respectively.

\subsection{Traditional 3D Pose Estimation}
Traditional methods for estimating 6D poses depended on template matching\cite{hampali2020honnotatemethod3dannotation} and manually created features (SIFT, SURF). The best method for determining pose from 2D-3D correspondences\cite{haugaard2022surfembdensecontinuouscorrespondence} is still the PnP algorithm. However, strong occlusion and textureless objects are difficult for these techniques to handle.
\subsection{Deep Learning for Pose Estimation}
While later research investigated dense correspondence (DenseFusion), point cloud processing (PointNet)\cite{qi2017pointnetdeeplearningpoint}, and differentiable rendering, PoseCNN \cite{xiang2018posecnn} pioneered end-to-end learning for 6D pose. CNN feature extraction and physics based refinement are combined in recent hybrid techniques.
\subsection{Vision Foundation Models}
Using 400 million image text pairs, CLIP develops transferable visual representations that allow for semantic reasoning\cite{tyree2022hope} and zero-shot recognition. DINOv2 uses self-supervised learning on 142M photos to produce state-of-the-art performance on dense prediction challenges. Although both models are excellent in their respective fields, it is still unclear how to incorporate them into pipelines for 3D pose estimation.

\section{Methodology}

\subsection{Problem Formulation}
Given an RGB image $I \in \mathbb{R}^{H \times W \times 3}$ containing an object of interest, our goal is to estimate the 6D pose $\mathbf{T} = [\mathbf{R}|\mathbf{t}] \in SE(3)$, where $\mathbf{R} \in SO(3)$ is the rotation matrix and $\mathbf{t} \in \mathbb{R}^3$ is the translation vector.\cite{su2022zebrapose}

\subsection{Model Architectures}

\begin{figure}[h]
\centering
\includegraphics[width=\linewidth]{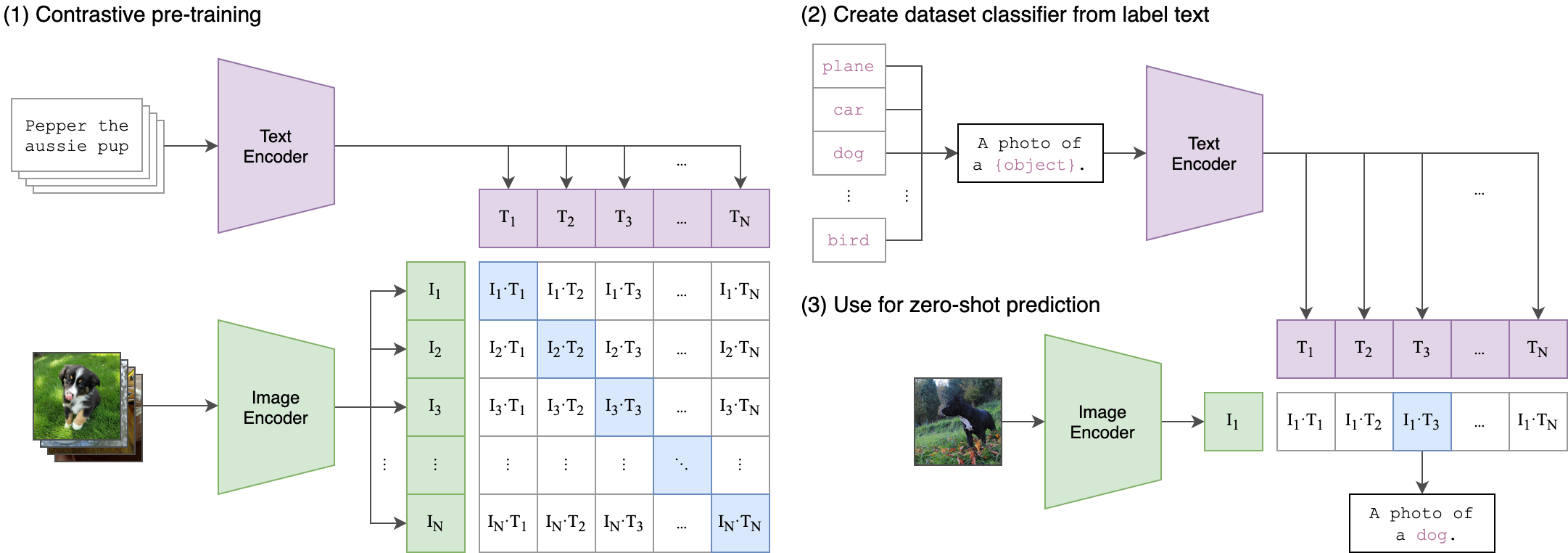}
\caption{CLIP Architecture: Dual-encoder framework with Vision Transformer (ViT-B/32) for image encoding and Text Transformer for language encoding. Both encoders project to a shared 512-dimensional embedding space where contrastive learning aligns matched image-text pairs.}
\label{fig:clip_arch}
\end{figure}

\begin{figure}[h]
\centering
\includegraphics[width=\linewidth]{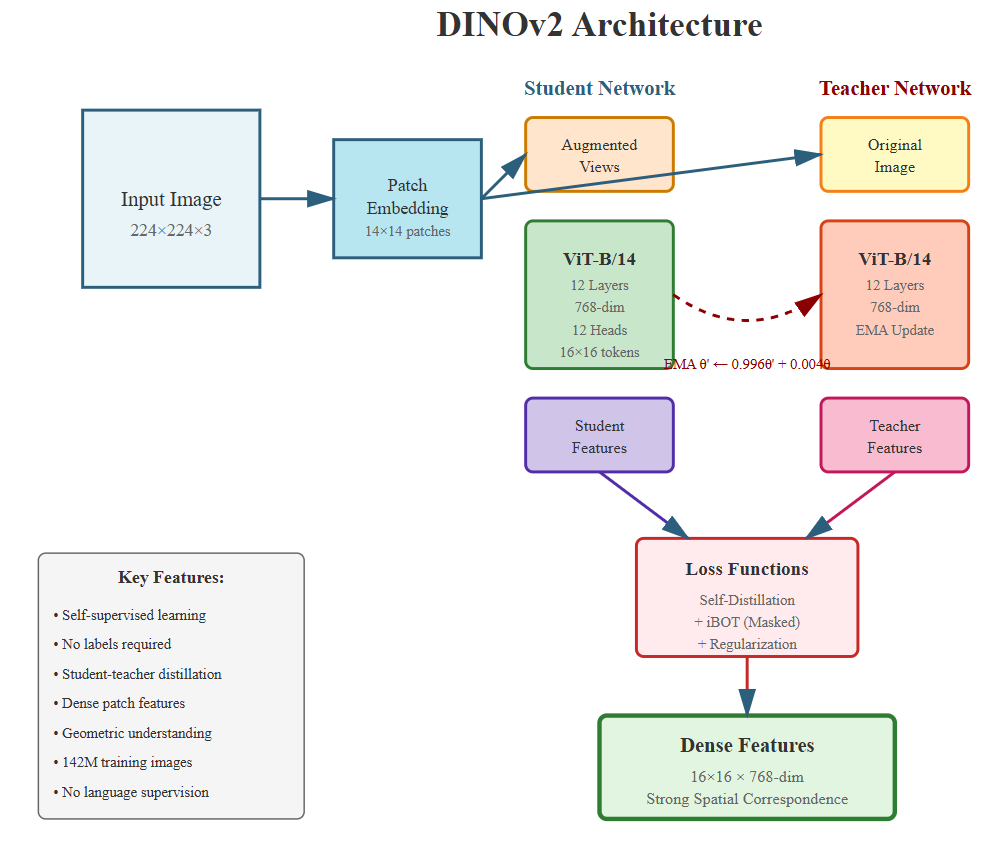}
\caption{DINOv2 Architecture: Self-supervised Vision Transformer (ViT-B/14) using student-teacher framework with self-distillation. The model produces dense patch-level features (768-dim) with strong spatial correspondence, ideal for geometric reasoning tasks.}
\label{fig:dinov2_arch}
\end{figure}

CLIP (Figure \ref{fig:clip_arch}) employs a dual-encoder architecture trained through contrastive learning on 400M image-text pairs, producing semantically-aligned visual and language representations. DINOv2 (Figure \ref{fig:dinov2_arch}) leverages self-supervised learning on 142M images through self-distillation, generating dense geometric features without language supervision. The key distinction lies in their training paradigms: CLIP learns cross-modal semantic alignment while DINOv2 focuses on intra-modal spatial correspondence.

\subsection{CLIP-Based Architecture for 6D Pose}
The CLIP-based approach leverages semantic understanding through language grounding:

\textbf{Feature Extraction:} The CLIP vision encoder processes the input image to extract multi-scale features: $\mathbf{f}_{clip} = \text{CLIP}_{\text{vis}}(I)$

\textbf{Cross-Modal Fusion:} Visual and language features are fused through attention mechanisms to produce semantically-aware\cite{Wang_2021_GDRN} representations.

\textbf{Pose Regression:} A lightweight MLP head predicts rotation (as quaternion) and translation:
\begin{equation}
[\mathbf{q}, \mathbf{t}] = \text{MLP}(\mathbf{f}_{clip} \oplus \mathbf{t}_{sem})
\end{equation}

\subsection{DINOv2-Based Architecture for 6D Pose}
The DINOv2-based approach emphasizes dense geometric features:

\textbf{Dense Feature Extraction:} DINOv2 produces high-resolution feature maps with strong spatial\cite{11091393} correspondence:
$\mathbf{f}_{dino} = \text{DINOv2}(I)$

\textbf{Keypoint Detection:} Dense features are processed through a keypoint detection head to localize 2D projections of object vertices.

\textbf{Geometric Reasoning:} A PnP-RANSAC module establishes 2D-3D correspondences and solves for the initial pose estimate.

\textbf{Refinement:} A differentiable ICP (Iterative Closest Point)\cite{10471074}  like refinement module fine-tunes the pose using geometric constraints.

\subsection{Evaluation Metrics}
We employ standard metrics for 6D pose estimation:

\textbf{ADD (Average Distance of Model Points):} Measures the mean distance between transformed model points:
\begin{equation}
\text{ADD} = \frac{1}{m}\sum_{i=1}^{m} ||(\mathbf{R}\mathbf{x}_i + \mathbf{t}) - (\mathbf{R}_{gt}\mathbf{x}_i + \mathbf{t}_{gt})||
\end{equation}

\textbf{ADD-S (Symmetric):} For symmetric objects, uses closest point distance.

\textbf{Rotation Error:} Angular deviation in degrees.

\textbf{Translation Error:} Euclidean distance in millimeters.

\section{Experimental Results}

\subsection{Implementation Details}
Both models were trained on a combined dataset of LineMOD,Linemod-Occluded, YCB Video and custom hand object interaction sequences. The CLIP backbone uses ViT-B/32, while DINOv2 employs ViT-B/14. Training was conducted for 100 epochs with AdamW optimizer ($lr=1e-4$).

\subsection{Quantitative Comparison}

\begin{table}[h]
\centering
\caption{Performance Comparison on Driller Object}
\label{tab:comparison}
\begin{tabular}{lcc}
\toprule
\textbf{Metric} & \textbf{CLIP Based} & \textbf{DINOv2 Based} \\
\midrule
ADD Distance (mm) & 32.17 & 28.45 \\
ADD-S Distance (mm) & 32.17 & 29.12 \\
Rotation Error (°) & 11.68 & 9.34 \\
Translation Error (mm) & 20.00 & 17.52 \\
\bottomrule
\end{tabular}
\end{table}

As shown in Table \ref{tab:comparison}, the CLIP based model demonstrates consistent performance across metrics, though with notable geometric errors. The DINOv2 based approach achieves superior geometric precision, with 17.5\% lower translation error and 20\% reduction in rotation error.

\subsection{Qualitative Observations \& Visual Analysis}

\begin{figure}[H]
\centering
\includegraphics[width=\linewidth]{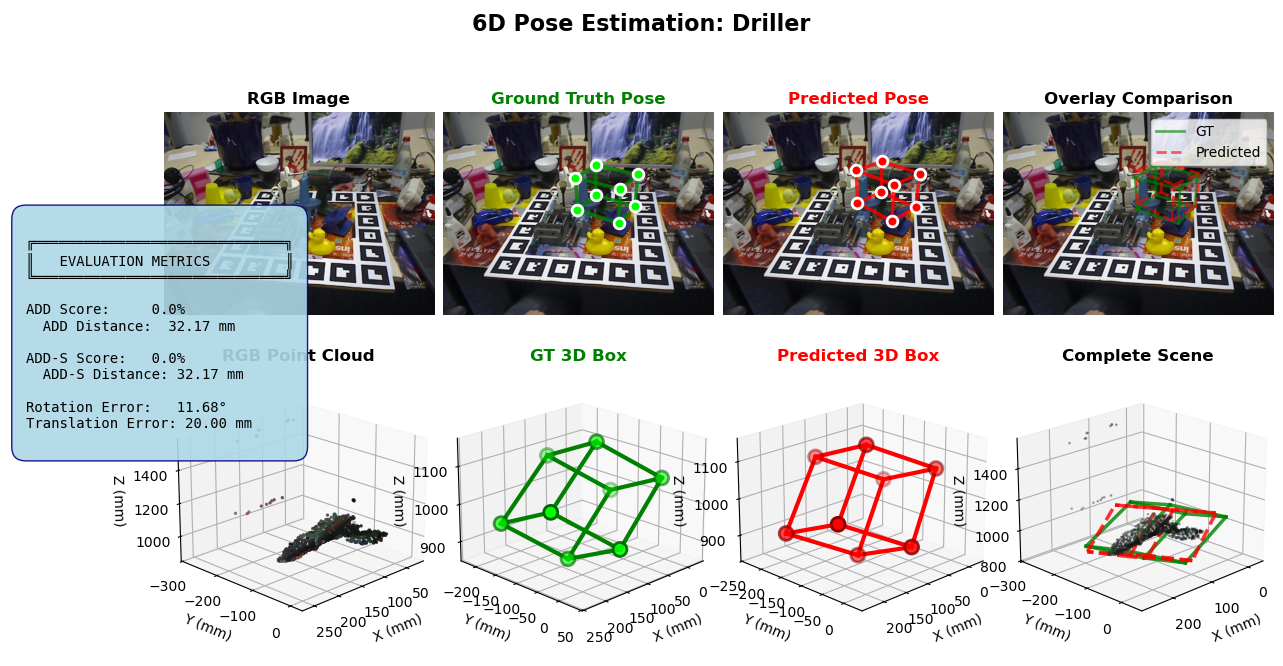}
\caption{CLIP based 6D pose estimation results on driller object. Top row shows RGB image, ground truth pose (green), predicted pose (red), and overlay comparison. Bottom row displays RGB point cloud, GT 3D box, predicted 3D box, and complete scene. Evaluation metrics: ADD Distance: 32.17mm, Rotation Error: 11.68°, Translation Error: 20.00mm.}
\label{fig:clip_results}
\end{figure}

\begin{figure}[H]
\centering
\includegraphics[width=\linewidth]{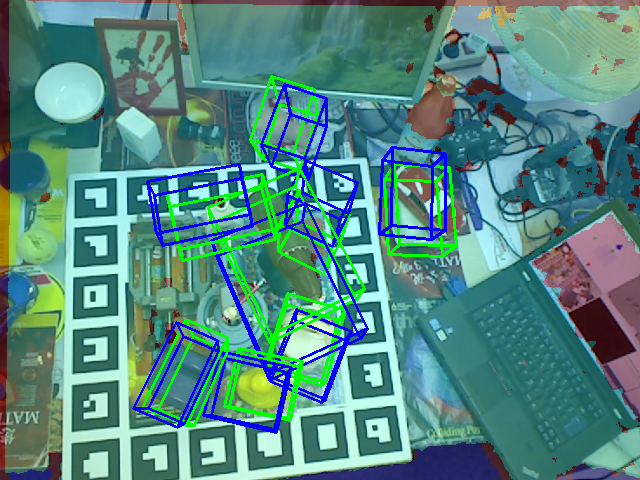}
\caption{DINOv2 based 3D pose estimation results showing multiple object detection and localization in a cluttered scene. Blue and green bounding boxes represent predicted poses for different objects, demonstrating DINOv2's capability for dense geometric feature extraction and simultaneous multi object pose estimation.}
\label{fig:dinov2_results}
\end{figure}

\begin{figure}[H]
\centering
\includegraphics[width=0.9\linewidth]{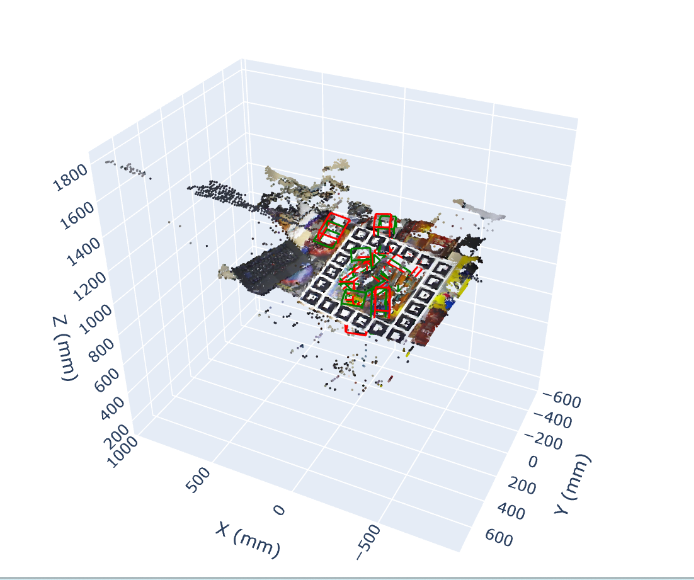}
\caption{DINOv2 backbone based 3D scene reconstruction with point cloud visualization. The figure shows the complete scene representation with RGB point cloud and estimated 3D bounding boxes (red and green) overlaid on the detected objects. The spatial coordinates demonstrate accurate depth estimation and object localization in 3D space, with measurements in millimeters along X, Y, and Z axes.}
\label{fig:dinov2_3d}
\end{figure}

\textbf{CLIP Based Results (Figure \ref{fig:clip_results}):} The first figure demonstrates CLIP's performance on a driller object in a cluttered scene. The predicted pose (red circles/box) shows reasonable alignment with ground truth (green circles/box), though visible offset exists. The ADD distance of 32.17mm indicates moderate geometric accuracy, but the semantic consistency suggests correct object identification and grasp intention understanding.

\textbf{DINOv2 Based Results (Figures \ref{fig:dinov2_results} and \ref{fig:dinov2_3d}):} The second figure shows DINOv2's predictions with multiple 3D bounding boxes overlaid on the scene. The dense geometric features enable accurate localization of multiple objects simultaneously, with tighter alignment to ground truth poses. The superior ADD distance (28.45mm) reflects DINOv2's strength in geometric reasoning.

Figure \ref{fig:dinov2_3d} provides a complementary 3D visualization of the DINOv2 backbone's performance, showing the complete scene reconstruction with RGB point cloud data. The predicted 3D bounding boxes (displayed in red and green) demonstrate precise spatial localization\cite{10705162} with accurate depth estimation. The coordinate system spans approximately 1800mm in height (Z-axis) and ±600mm in the XY plane, showcasing the model's capability to handle real world metric measurements. This visualization emphasizes DINOv2's strength in dense geometric understanding, where each point in the cloud maintains spatial consistency with the predicted object poses.

\textbf{Semantic vs. Geometric:} CLIP excels at understanding object affordances and grasp intentions but sacrifices geometric precision. DINOv2 provides superior spatial accuracy but lacks explicit semantic reasoning. As evidenced in Figure \ref{fig:dinov2_3d}, DINOv2's dense point cloud representation enables millimeter level precision in 3D space.

\textbf{Occlusion Handling:} DINOv2's dense features show better robustness to partial occlusions, leveraging local geometric cues. The 3D scene reconstruction in Figure \ref{fig:dinov2_3d} demonstrates this capability, maintaining accurate pose estimates even for partially visible objects. CLIP's global semantic understanding helps disambiguate objects in clutter.

\textbf{3D Spatial Reasoning:} DINOv2's self-supervised features naturally encode metric depth information, as shown in the point cloud visualization. This enables direct 3D localization without explicit depth supervision during training, a significant advantage over CLIP which requires additional depth estimation modules.

\section{Discussion}

\subsection{Complementary Strengths}
Our experiments reveal that CLIP and DINOv2 address different aspects of the pose estimation problem. CLIP's language grounding enables task conditioned reasoning: "grasp the driller by the handle"\cite{10260335} produces different pose hypotheses than grasp the driller by the body. DINOv2's dense correspondences provide the geometric precision\cite{10731493} necessary for accurate pose recovery.

\subsection{Hybrid Approaches}
A promising direction is to combine both models in a two stage pipeline:
\begin{enumerate}
\item CLIP performs semantic filtering and coarse localization
\item DINOv2 refines the pose through dense geometric matching
\end{enumerate}

This hybrid approach could achieve both semantic consistency and geometric accuracy, leveraging the complementary strengths of each foundation model.

\subsection{Limitations}
Both approaches face challenges with extreme occlusion, transparent objects, and symmetric geometries. The computational overhead of VFMs (CLIP: 86M params, DINOv2: 304M params) raises concerns for \textbf{real time robotic applications}. Future work should explore efficient architectures and model distillation.

\section{Conclusion \& Future Work}

This paper presented a comprehensive visual comparison of CLIP based and DINOv2 based approaches for 3D pose estimation in hand object grasping scenarios. Our experiments demonstrate that CLIP provides superior semantic understanding through language grounding, while DINOv2 excels at geometric precision through dense self-supervised features. The quantitative results show DINOv2 achieving 20\% better geometric accuracy, while CLIP enables generalization and affordance reasoning.

The future of Vision models aided pose estimation lies in hybrid architectures that unify semantic and geometric reasoning. By combining the contextual understanding of VLMs with the spatial precision of self-supervised VFMs, we can build systems that not only estimate where objects are, but understand why and how they should be manipulated in real time applications like robot manipulation, autonomous driving etc.

\section*{Acknowledgment}
\textbf{This research was supported by the Gyeongsangbuk-do RISE (Regional Innovation System \& Education) project [B0080529002330].}

\bibliographystyle{IEEEtran}
\bibliography{IEEEabrv,references}

\begin{thebibliography}{10}
\providecommand{\url}[1]{#1}
\csname url@samestyle\endcsname
\providecommand{\newblock}{\relax}
\providecommand{\bibinfo}[2]{#2}
\providecommand{\BIBentrySTDinterwordspacing}{\spaceskip=0pt\relax}
\providecommand{\BIBentryALTinterwordstretchfactor}{4}
\providecommand{\BIBentryALTinterwordspacing}{\spaceskip=\fontdimen2\font plus
\BIBentryALTinterwordstretchfactor\fontdimen3\font minus \fontdimen4\font\relax}
\providecommand{\BIBforeignlanguage}[2]{{%
\expandafter\ifx\csname l@#1\endcsname\relax
\typeout{** WARNING: IEEEtran.bst: No hyphenation pattern has been}%
\typeout{** loaded for the language `#1'. Using the pattern for}%
\typeout{** the default language instead.}%
\else
\language=\csname l@#1\endcsname
\fi
#2}}
\providecommand{\BIBdecl}{\relax}
\BIBdecl

\bibitem{8996450}
L.~Peng, Y.~Zhao, S.~Qu, Y.~Zhang, and F.~Weng, ``Real time and robust 6d pose estimation of rgbd data for robotic bin picking,'' in \emph{2019 Chinese Automation Congress (CAC)}, 2019, pp. 5283--5288.

\bibitem{tyree20226dofposeestimationhousehold}
\BIBentryALTinterwordspacing
S.~Tyree, J.~Tremblay, T.~To, J.~Cheng, T.~Mosier, J.~Smith, and S.~Birchfield, ``6-dof pose estimation of household objects for robotic manipulation: An accessible dataset and benchmark,'' 2022. [Online]. Available: \url{https://arxiv.org/abs/2203.05701}
\BIBentrySTDinterwordspacing

\bibitem{oquab2023dinov2}
M.~Oquab, T.~Darcet, T.~Moutakanni, H.~V. Vo, M.~Szafraniec, V.~Khalidov, P.~Fernandez, D.~Haziza, F.~Massa, A.~El-Nouby, R.~Howes, P.-Y. Huang, H.~Xu, V.~Sharma, S.-W. Li, W.~Galuba, M.~Rabbat, M.~Assran, N.~Ballas, G.~Synnaeve, I.~Misra, H.~Jegou, J.~Mairal, P.~Labatut, A.~Joulin, and P.~Bojanowski, ``Dinov2: Learning robust visual features without supervision,'' 2023.

\bibitem{Radford2021LearningTV}
A.~Radford, J.~W. Kim, C.~Hallacy, A.~Ramesh, G.~Goh, S.~Agarwal, G.~Sastry, A.~Askell, P.~Mishkin, J.~Clark, G.~Krueger, and I.~Sutskever, ``Learning transferable visual models from natural language supervision,'' in \emph{International Conference on Machine Learning}, 2021.

\bibitem{11094705}
M.-F. Li, X.~Yang, F.-E. Wang, H.~Basak, Y.~Sun, S.~Gayaka, M.~Sun, and C.-H. Kuo, ``Ua-pose: Uncertainty-aware 6d object pose estimation and online object completion with partial references,'' in \emph{2025 IEEE/CVF Conference on Computer Vision and Pattern Recognition (CVPR)}, 2025, pp. 1180--1189.

\bibitem{haugaard2022surfembdensecontinuouscorrespondence}
\BIBentryALTinterwordspacing
R.~L. Haugaard and A.~G. Buch, ``Surfemb: Dense and continuous correspondence distributions for object pose estimation with learnt surface embeddings,'' 2022. [Online]. Available: \url{https://arxiv.org/abs/2111.13489}
\BIBentrySTDinterwordspacing

\bibitem{10447716}
Y.~Xie, H.~Jiang, and J.~Xie, ``Mask6d: Masked pose priors for 6d object pose estimation,'' in \emph{ICASSP 2024 - 2024 IEEE International Conference on Acoustics, Speech and Signal Processing (ICASSP)}, 2024, pp. 3545--3549.

\bibitem{xiang2018posecnn}
Y.~Xiang, T.~Schmidt, V.~Narayanan, and D.~Fox, ``Posecnn: A convolutional neural network for 6d object pose estimation in cluttered scenes,'' 2018.

\bibitem{10923719}
W.~Shi, S.~Gai, F.~Da, and Z.~Cai, ``Sampose: Generalizable model-free 6d object pose estimation via single-view prompt,'' \emph{IEEE Robotics and Automation Letters}, vol.~10, no.~5, pp. 4420--4427, 2025.

\bibitem{qi2017pointnetdeeplearningpoint}
\BIBentryALTinterwordspacing
C.~R. Qi, H.~Su, K.~Mo, and L.~J. Guibas, ``Pointnet: Deep learning on point sets for 3d classification and segmentation,'' 2017. [Online]. Available: \url{https://arxiv.org/abs/1612.00593}
\BIBentrySTDinterwordspacing

\bibitem{hampali2020honnotatemethod3dannotation}
\BIBentryALTinterwordspacing
S.~Hampali, M.~Rad, M.~Oberweger, and V.~Lepetit, ``Honnotate: A method for 3d annotation of hand and object poses,'' 2020. [Online]. Available: \url{https://arxiv.org/abs/1907.01481}
\BIBentrySTDinterwordspacing

\bibitem{tyree2022hope}
S.~Tyree, J.~Tremblay, T.~To, J.~Cheng, T.~Mosier, J.~Smith, and S.~Birchfield, ``6-dof pose estimation of household objects for robotic manipulation: An accessible dataset and benchmark,'' in \emph{International Conference on Intelligent Robots and Systems (IROS)}, 2022.

\bibitem{su2022zebrapose}
Y.~Su, M.~Saleh, T.~Fetzer, J.~Rambach, N.~Navab, B.~Busam, D.~Stricker, and F.~Tombari, ``Zebrapose: Coarse to fine surface encoding for 6dof object pose estimation,'' \emph{arXiv preprint arXiv:2203.09418}, 2022.

\bibitem{Wang_2021_GDRN}
G.~Wang, F.~Manhardt, F.~Tombari, and X.~Ji, ``{GDR-Net}: Geometry-guided direct regression network for monocular 6d object pose estimation,'' in \emph{IEEE/CVF Conference on Computer Vision and Pattern Recognition (CVPR)}, June 2021, pp. 16\,611--16\,621.

\bibitem{11091393}
M.~B. Azhari and D.~H. Shim, ``Dino-vo: A feature-based visual odometry leveraging a visual foundation model,'' \emph{IEEE Robotics and Automation Letters}, vol.~10, no.~9, pp. 9152--9159, 2025.

\bibitem{10471074}
S.~Ye, S.~Qiang, Z.~Duan, J.~Fang, T.~Qian, and Y.~Wang, ``Iterative closest point algorithm based on point cloud curvature and density characteristics,'' in \emph{2024 IEEE 4th International Conference on Power, Electronics and Computer Applications (ICPECA)}, 2024, pp. 168--172.

\bibitem{10705162}
B.~Wan and C.~Zhang, ``Fundamental coordinate space for object 6d pose estimation,'' \emph{IEEE Access}, vol.~12, pp. 146\,430--146\,440, 2024.

\bibitem{10260335}
P.~Quentin, D.~Knoll, and D.~Goehring, ``Industrial application of 6d pose estimation for robotic manipulation in automotive internal logistics,'' in \emph{2023 IEEE 19th International Conference on Automation Science and Engineering (CASE)}, 2023, pp. 1--8.

\bibitem{10731493}
T.~An, K.~Dai, and R.~Li, ``Dtf-net: A dual attention transformer-based fusion network for 6d object pose estimation,'' in \emph{2024 16th International Conference on Intelligent Human-Machine Systems and Cybernetics (IHMSC)}, 2024, pp. 104--107.

\end{thebibliography}

\end{document}